# Interior Object Geometry via Fitted Frames[1]

Stephen M. Pizer, Zhiyuan Liu, Junjie Zhao, Nicholas Tapp-Hughes, James Damon[2], Miaomiao Zhang***, JS Marron, Mohsen Taheri**, Jared Vicory*
University of North Carolina at Chapel Hill, Chapel Hill, NC 27514 USA;
*Kitware, Inc., Carrboro, NC USA; **University of Stavanger; ***University of Virginia
Corresponding author: SM Pizer, pizer@cs.unc.edu

## Abstract

We propose a means of computing fitted frames on the boundary *and in the interior of objects* and using them to provide the basis for producing geometric features from them that are not only alignment-free but most importantly can be made to correspond locally across a population of objects. We describe a representation targeted for anatomic objects which is designed to enable this strong locational correspondence within object populations and thus to provide powerful object statistics. It accomplishes this by understanding an object as the diffeomorphic deformation of the closure of the interior of an ellipsoid and by using a skeletal representation fitted throughout the deformation to produce a model of the target object, where the object is provided initially in the form of a boundary mesh. Via classification performance on hippocampi shape between individuals with a disorder vs. others, we compare our method to two state-of-the-art methods for producing object representations that are intended to capture geometric correspondence across a population of objects and to yield geometric features useful for statistics, and we show notably improved classification performance by this new representation, which we call the *evolutionary s-rep*. The geometric features that are derived from each of the representations, especially via fitted frames, are discussed.

## 1. Introduction

If one wants to do statistics, such as classification, hypothesis testing, or calculation of means, on object shape populations, such the one shown in Fig. 1 middle, it is important that the features used reflect as much as possible agreement on the localization of geometric properties held in common across the population. This characteristic is called *correspondence*. We compare two categories of object representations used for producing such object correspondences, given smooth-boundaried object samples in the training and test sets specified by a relatively dense boundary mesh. Of the 3D object representations that have been claimed to be promising for statistics, the two most promising categories are the ones we study:

1. Diffeomorphisms over 3D space derived via an LDDMM algorithm from the object mesh vertices [Durrleman 2014] or from the binary images describing the objects [Zhang 2019]; these are represented respectively by momentum images or initial velocity images, i.e., 3D arrays of vectors, with the 3D space covered by the array containing the objects.

2. The skeletal representation called evolutionary s-reps described in this paper. More history and details of s-reps are given in [Pizer 2022]; in 3D these are represented by a

[1] This paper is associated with a companion paper: Vicory et al., Fitting, Extraction, and Evaluation of Geometric Features To Produce Correspondence for Statistics
[2] Posthumously

2D spatial skeletal grid in which at each grid-point a collection of directions and lengths from the skeleton to the boundary are provided (Fig. 4). Novel here is using the s-rep to produce fitted-frame-based curvatures and inter-grid-point lengths in the closure of the object interior, where these features are created in locational correspondence using evolution from an ellipsoid based on rich, s-rep-based geometric properties.

The objects of our interest are 3D anatomic objects. These are smooth-boundaried, and each object maintains its topology over its variations (see Fig. 1). For the sake of simplicity, in this paper we restrict ourselves to objects of spherical topology, but other topologies, such as toroidal, also can be dealt with by our methods. Many of the objects are "slabular', that is, [Damon 2021] they can be swept by a set of smooth 2D surfaces that do not intersect within the object with the intersecting 2D surfaces having consistent topology (Fig. 1 right). We treat objects with a circular cross-section as a special case, not covered here, and, as described next, we treat notable protrusions or indentations by separately representing them and the connections to a host figure (Fig. 2). Here we only discuss the representation of host figures that have no separable protrusions or indentations.

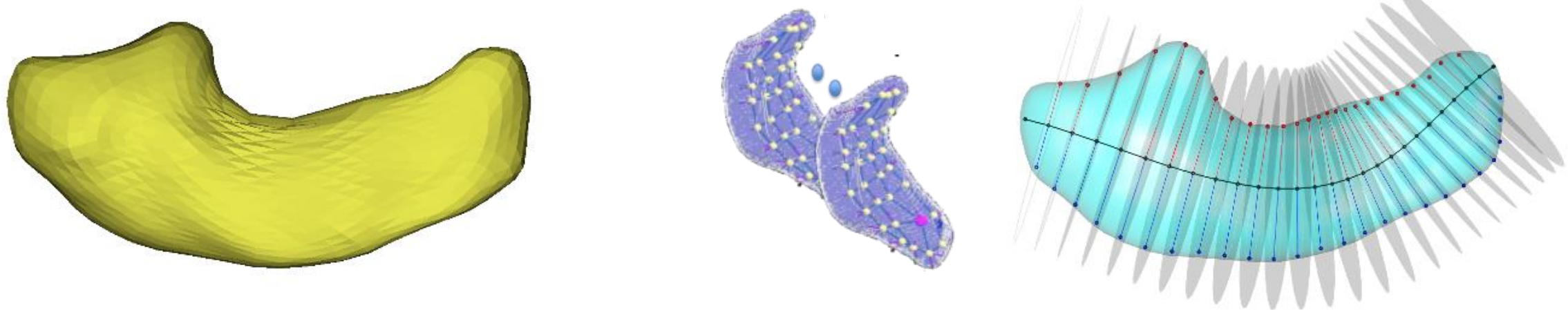

Fig. 1. Left: a hippocampus.  Middle: a population of hippocampi. Right. The hippocampus being swept by planes not intersecting within the object

Our treatment of host figures occurs via the example of the mandible (Fig. 2). It can be seen as a non-slabular object that has protrusions from a host slabular surface. These protrusions are formed by the teeth and the anterior horns where the mandible intersects the skull; these are called the coronoid processes. The host can be obtained by a sort of smoothing of the object surface, e.g., by conformalized mean curvature flow [Kazhdan 2012]. Representing such slabular objects can be accomplished by the method described in this paper.

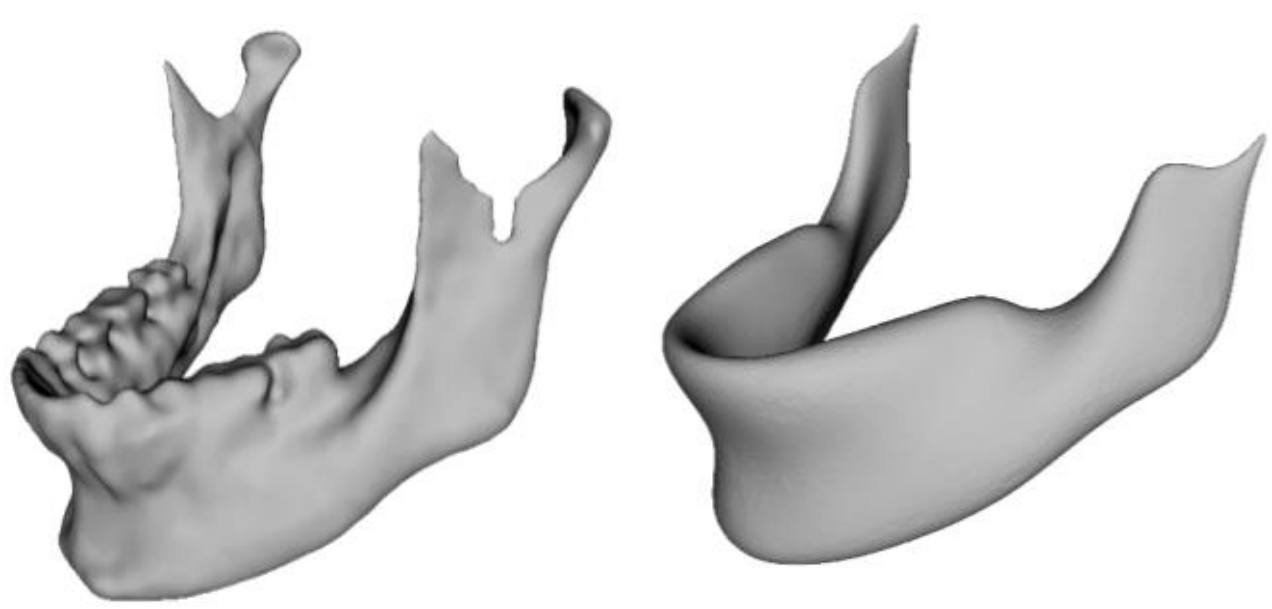

Figure 2. Left: a mandible, which is non-slabular before it was smoothed into the object on the right, which can serve as the host of its protrusions.

In this paper we focus on the theoretical and conceptual aspect of our approach. Our companion paper [Vicory 2025] focuses on the experimental and computational aspects that allow for derivation of our s-reps and the diffeomorphisms to be derived from input object boundary meshes, for the geometric features to be calculated from the geometric representations, and for the comparisons via classification performance to be achieved.

Achieving statistical correspondence over a population of objects has been approached in two ways:

A. Defining correspondence via an information-theoretic measure such as entropy on the basic boundary representation and optimizing that measure for the given training population [Cootes 2001; Cates 2007]. Due to the large number of boundary features per object and the large number of objects in the collection of objects used to train the statistics, this approach has been found unwieldy and does not always achieve a global optimum.

B. Using a form of representation seen to richly reflect object geometric properties, to fit this to members of the object population, to build the statistical approach of interest based on the derived features, and to empirically determine the quality of the representation by measuring the statistical performance using those features.

This paper uses this latter approach on the two geometric representations listed above: diffeomorphisms' momenta or initial velocities and s-reps features, with the s-reps approach having two variants: one produced in the work of Z. Liu [2021], in which the deformation from an ellipsoid to a target object is directly computed, and the evolutionary method described here, in which the fitting occurs throughout the evolution from the ellipsoid to the target object. The statistical strength of each of the three is measured by classification performance on infants' hippocampi discriminated into autistic and control classes.

Early intuition suggested that the boundary point distribution model (PDM), perhaps with sparser spatial sampling than in the input form of the objects, perhaps with correspondence provided via spherical harmonics [Styner 2006] might provide a fully adequate form for statistical analysis. However, various work measuring effectiveness via various statistical properties [e.g., Hong 2016, Tu 2016, Liu 2021, Pizer & Marron 2017] has suggested that this is inferior to the two derived representations listed earlier, which are designed to capture a richer set of geometric properties than boundary locations alone.

The LDDMM technique uses the input boundary mesh points or binary image to first produce an alignment over the training and test cases and then from each object's results to derive a diffeomorphism of 3D space from their mean to the object that is represented by a relatively dense 3D array of vectors called a "momentum" or "initial velocities". Such an array of velocities specifies a geodesic in the space of diffeomorphisms and can be understood as a collection of geometric properties on the tangent hyperplane to the identity diffeomorphism in the curved manifold of diffeomorphisms. While somewhat commonly the momentum or initial velocity vectors are represented as $(x,y,z)$ tuples, because of the importance of direction as a feature, we represent each vector as a length and a unit vector indicating displacement direction of the associated 3-space point.

Both the PDM approach and the LDDMM approach depend on pre-alignment, which propagates its error into the determination of correspondence. Also, they both ignore geometry related to the object interior, such as a) the curvature information of the 2D object shown in Fig. 3 and b) the fact that that object is understood to have a bulge in the middle. Both the curvature and width features can be understood by fitting the boundary with a skeletal axis equipped with spoke vectors from the axis to the boundary. Moreover, we will show how the skeleton and its spokes can be used to avoid pre-alignment and to capture interior geometric features.

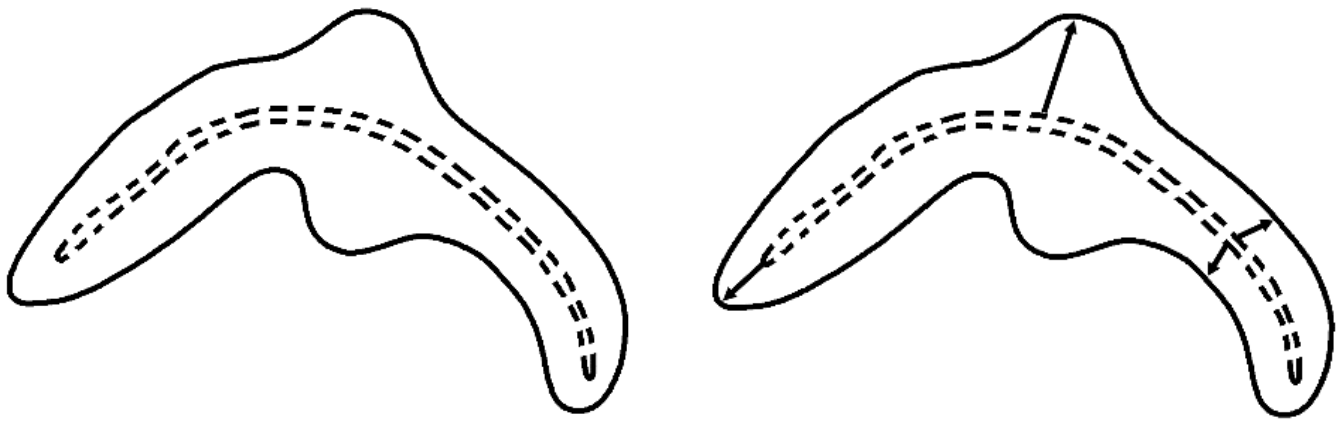

Figure 3: Left: 2D object with curvature and a bulge, showing as dashed its skeletal (interior) axis (both dashed sides are actually co-located). Right: “Spoke” vectors from the skeleton to the boundary capturing object width. Together the skeleton and the spokes form an s-rep. Fig. 4 shows a similar representation, called an “s-rep” for a 3D object.

The s-reps approach is built on the intuition that object width and curvature of the object interior as they vary across the object are especially indicative of object shape and the way in which it corresponds across a population. According to [Damon 2008a] for a 2D continuous object, the object boundary can be understood as the dilation of a surface, called a skeleton, made of two co-located copies of a curve that folds back onto itself at a so-called fold point when it turns from the top side to the bottom side by a turn of angle $\pi$. The dilation occurs in proportion to the spoke vectors, which connect the skeleton to the boundary without crossing. Similarly, for a 3D object the skeleton is two co-located copies of a curved surface that folds back onto itself by a turn of angle $\pi$ along a “fold” space curve (Fig. 4 bottom). Here again the dilation occurs in proportion to spoke vectors.

This spoke-proportional dilation exposes level surfaces (“onion skins”) of the dilation proportion, which in turn allow geometric properties of the object interior to be derived. This capture of features of the interior geometry over and above those of the boundary geometry is an important aspect of our method.

As described in [Pizer 2022], the discrete s-rep (which will be referred to as just the s-rep in what follows) consists of a sampled, folded skeletal surface and non-crossing spoke vectors from those skeletal points to the object boundary (see Fig. 4). In 3D the curvilinear skeleton of the skeleton surface is called the “spine”. In our recent work the s-rep is produced by computing, for any training or test object, a diffeomorphism of the closure of the object interior, alone, from a common ellipsoid (in 2D an ellipse) and applying that diffeomorphism to the s-rep of the ellipsoid. That ellipsoid is seen as the most basic form of shape; its Blum medial skeleton is analytically known to be a folded planar ellipse. That ellipse can be understood to have its own skeleton, a folded line segment that we call its spine; that ellipse has its own medially described planar spokes, which we call “veins” (Fig. 4 bottom).

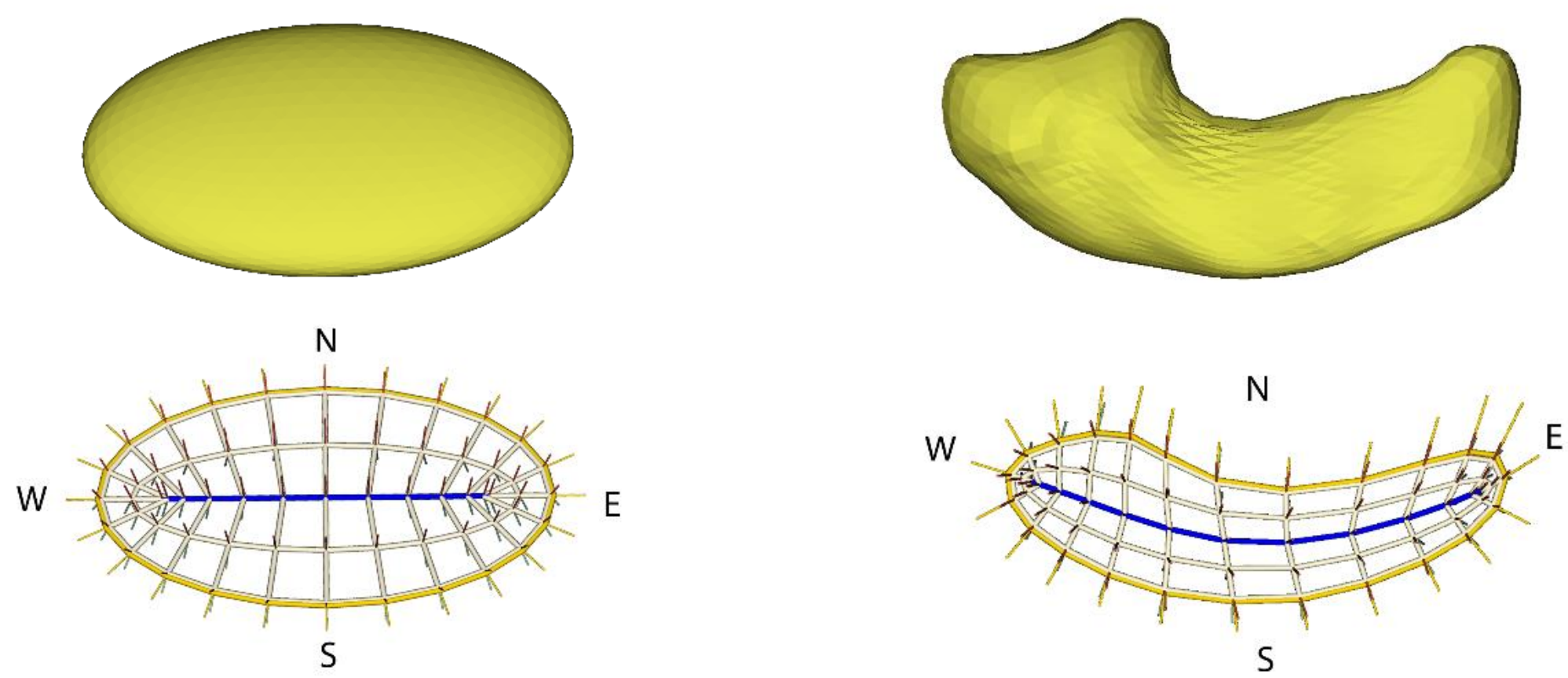

Figure 4. S-reps (on bottom row), shown from the top side, for an ellipsoid (left) and a hippocampus (right), with the top row showing the boundary. The s-reps’ spines are shown in bold blue, and shown in yellow are the fold curves, where the skeleton top folds over to the co-located skeletal bottom. Beige curvilinear segments show the grid on the skeleton in the representation; among these the curvilinear segments called “veins” proceed from the spine to both the north and south halves of the fold, which are separated by vertices of the fold curve at its “West” and “East” ends. For the ellipsoid the slightly oblique view shows not only the top-side spokes but bits of the ends of some of the bottom-side spokes.

The s-rep must be fitted to the input boundary in a way reflecting the curvature and width features of the object. The first method for doing this via an ellipsoid-based diffeomorphism was described by Z. Liu [2021] It involves finding an LDDMM deformation of the ellipsoid boundary to the target boundary, applying that to the ellipsoid s-rep, and refining the result to be closer to having Blum medial properties. However, our late colleague, James Damon, pointed out that the diffeomorphism produced by that method failed to reflect a variety of important geometric properties, leading to poor correspondence across the population: it did not map the ellipsoid s-rep to the target object s-rep in such a way that the spoke vectors remained straight nor that the so-called radial distances (fractions of the distance along spokes) mapped onto the same radial distances in the target object’s s-rep. Especially important were its following geometric failures:

1) It does not map the vertices (local maxima of Gaussian curvature) of the ellipsoid, which correspond to the vertices of the skeleton, onto ones of the target object.
2) It does not map the crests (space curves of maximal curvature along principal directions) of the ellipsoid, which correspond to the fold curves of the skeleton, onto crests of the target object.
3) It does not map the veins along the medial ellipse onto the corresponding curves of the target object’s skeleton (not implemented in the present program).

As such, the geometric properties of the resulting target s-reps for the training population would be expected to lead to poorer correspondence than desirable. This has led us to our new

"evolutionary" method producing an ellipsoid-based diffeomorphism that maintains these properties all the way through the application of the stages of the diffeomorphism.

Section 2 describes the major mathematical contribution of this paper: the means of deriving fitted frames in the closure of a target object's interior. These fitted frames (see Fig. 5) allow alignment-independent geometric features to be derived by using Elie Cartan's approach [1910] of representing curvatures and vectors in the coordinates of the local frame. Also, as with Cartan they also allow curvatures to be computed, but here not only on the boundary, which Cartan treated, but novelly also in the object interior. In section 2 you will see that the fitted frames depend on the spokes of the s-rep, more specifically on the geometry of the onion skins produced by dilating the skeleton by a proportion of the spoke length. Section 2 begins with some further description of s-reps.

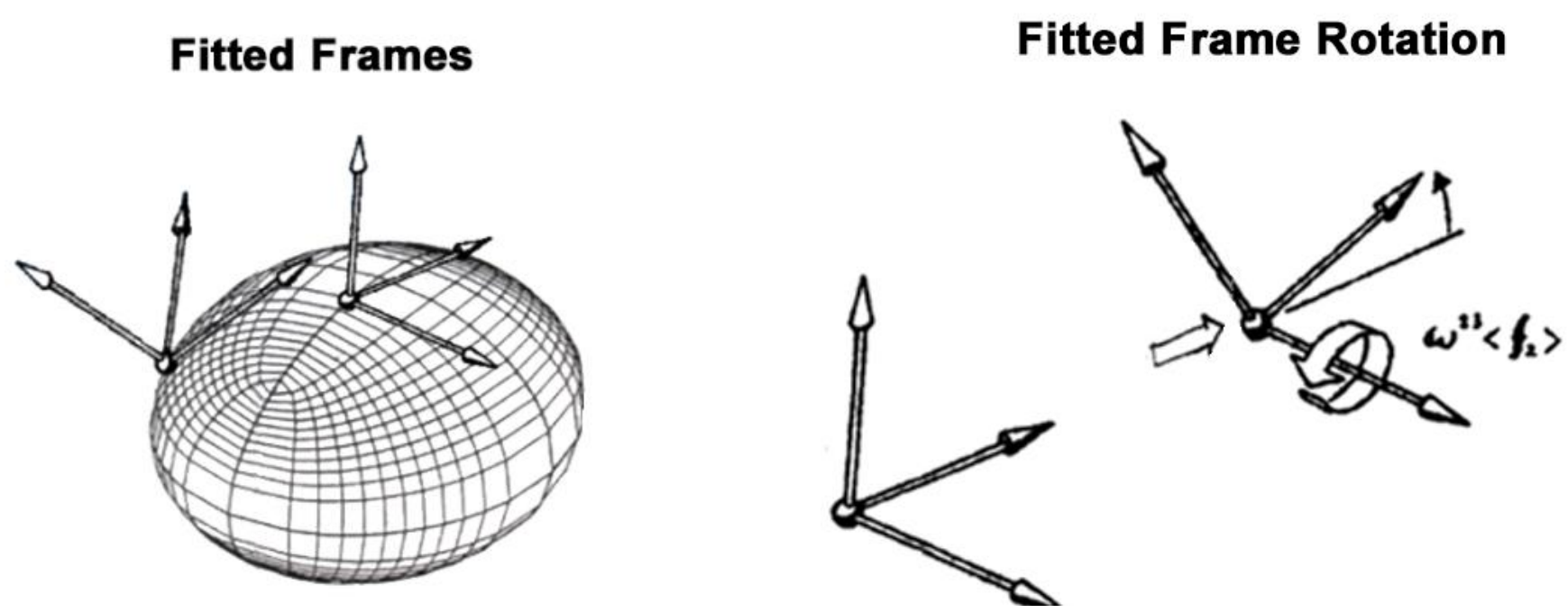

Fig. 5. Left: Fitted frames on two points of an ellipsoid boundary.
Right [from Koenderink 1990]: Rotation of a fitted frame in its own local coordinate system.

Section 3 shows how geometric features that are alignment-independent are derived using the fitted frames from the s-rep object models*[3]. After Section 4 briefly describes the LDDMM method for computing inter-object deformations*, a method used both in each stage of s-rep fitting and as a common method for statistical shape analysis, Section 5 overviews the method of fitting an s-rep to each of the objects in a sample set. Section 6 focuses on the results of our comparison experiment after overviewing the classification method we used for comparing the four object modeling approaches, the data that we used for the comparison, the geometric features derived from the LDDMM deformations, and the method for judging classification performance*. Section 7 discusses how s-reps can guarantee that the mean of a population has no local self-intersections. Section 8 reviews the approach of having fitted frames in the closure of an object's interior, presents the relative strengths of the methods we are comparing, and discusses the lessons on producing correspondence that were learned in this work and the limitations of our conclusions. Section 9 discusses various extensions of our correspondence-producing evolutionary s-reps approach. It describes the challenges ahead for making the s-rep based method of providing correspondence work on geometric entities of more complicated shape as well as ways of making s-rep fitting speedy.

[3] Details of all of the materials marked by an asterisk are covered in the companion paper [Vicory 2024].

## 2. Interior Frames from Object Boundaries for Statistical Analysis of Skeletons

Because pre-alignment of the objects in the population pollute the production of corresponding geometric features, we wish geometric features that are independent of the object alignment (rotation and translation). More importantly, we also want those features to richly characterize the object shape throughout the closure of its interior. To this end we describe a set of fitted frames computed at s-rep-associated discretely sampled points in that closure of the object interior, such that each geometric feature, at a point, is computed to be in the coordinate system provided by the fitted frame at that point. These s-rep-based fitted frames provide both of the aforementioned desired properties.

As shown in Fig. 4 and presented briefly earlier, a discrete skeleton of an s-rep for a slabular object consists of a grid of points on a surface folded onto itself that is smooth except at its fold such that it is dilated onto of the target object's boundary points. Associated with each skeletal point is a *spoke vector* to the boundary, such that the spokes fill the object interior and do not cross in that interior. That is, the skeleton is an entity with spherical topology that is folded in such a way that each non-fold point on its "top" side is co-located with a point on its "bottom" side. The spokes emanating from the fold connect to a crest point on the object, and the spokes from the vertices of the fold map onto vertices of the object. This behavior is important in making the s-rep richly represent the object's geometric properties. Section 5 shows how this behavior is accomplished for each object in the population.

Given that each object in the population has an s-rep in correspondence, this section lays out the important mathematical objective of producing fitted frames on the objects' interiors' closures, allowing geometric features to be based on their local fitted frames and thus not depend on pre-alignment. This capability of generating geometric features of curvature and inter-position distance by the use of local fitted frames was pioneered by mathematician Eli Cartan in his breakthrough 191- paper [Cartan 2010]. But those frames were produced only on object boundaries (Fig. 5). Here we show how to use the s-rep to produce frames in the object interior as well.

First, we demonstrate a set of s-rep-based intra-object coordinates[4] that lead to a correspondence that is insensitive to uniform widening 1) across the s-rep spokes, 2) along the long axis of the object, and 3) across the skeleton from one fold side to the other. The first of these coordinates is Damon's [Damon 2008a] *radial distance*, which is the fraction of the distance along the spoke from the skeleton to the boundary. We denote this coordinate $\boldsymbol{\tau_2}$; its value $\in [0,1]$, being 0.0 at the skeleton and 1.0 at the boundary. The second coordinate is cyclic along the spine of the s-rep, passing along the top side of the spine and back along the bottom side of the spine. We denote this coordinate $\boldsymbol{\theta}$, which we take to be 0 at the center of the spine and proceed E to W (see Fig. 6) counterclockwise on the top side to $\pi/2$ at one skeleton vertex (whose spoke touches a boundary vertex), then along the bottom side of the skeleton to $3\pi/2$ (=$-\pi/2$) at a second skeleton vertex, and then back along the top side to $2\pi$ (=0). The third coordinate captures distance from the spine along the veins as fraction of their length, together with a flag indicating which side of the spine the vein is. We denote this coordinate $\boldsymbol{\tau_1} \in [0,1]$. Veins are loci on the top side of the skeleton proceeding from the north side of the skeletal fold to the spine and then on to the south side. All of these coordinates are fractions, so they already are not sensitive to uniform scaling respectively in the principal length of the object (the spine length), in the cross-object (veins) widths, and in the short-axis widths (spoke lengths).

[4] These coordinates are specified slightly differently in [Pizer 2022].

Importantly, every position in the closure of the interior of an object has a unique value of $(\boldsymbol{\theta},\boldsymbol{\tau}_1,\boldsymbol{\tau}_2)$. Moreover, if the s-rep fitting is adequately reflective of the object geometry, this 3-tuple should be used as the basis for correspondence. Any position in one object that has a particular value of this tuple is taken to be in correspondence with that with the same tuple value in another object.

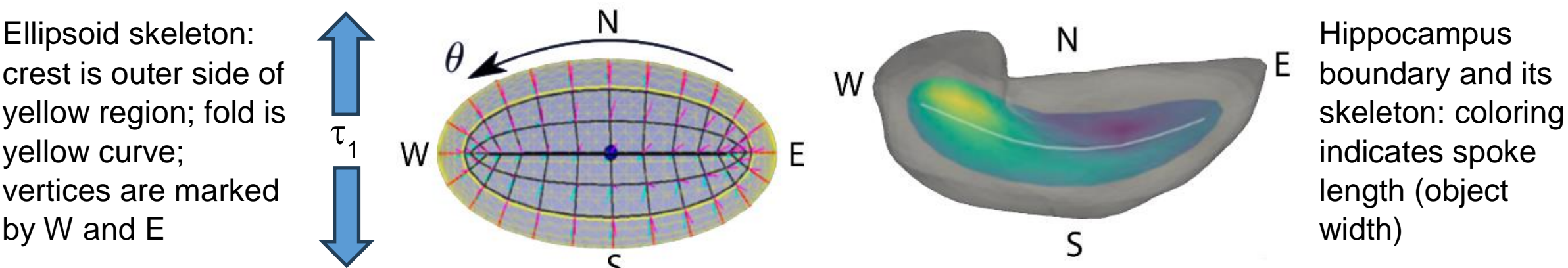

Fig. 6. Object coordinates for ellipsoid and hippocampus.

As Damon [2008a] described, level sets of the radial distance $\boldsymbol{\tau}_2$ form onion skins (Fig. 7). Our contribution is to let the geometry of the sequence of onion skins determine the fitted frames $\mathbf{F}(\boldsymbol{\theta},\boldsymbol{\tau}_1,\boldsymbol{\tau}_2)$ throughout the closure of the object interior (Fig. 8). Then the geometric features based on these frames are computed at a grid of points throughout the object, including on the skeleton, in the interior other than the skeleton, and on the boundary.

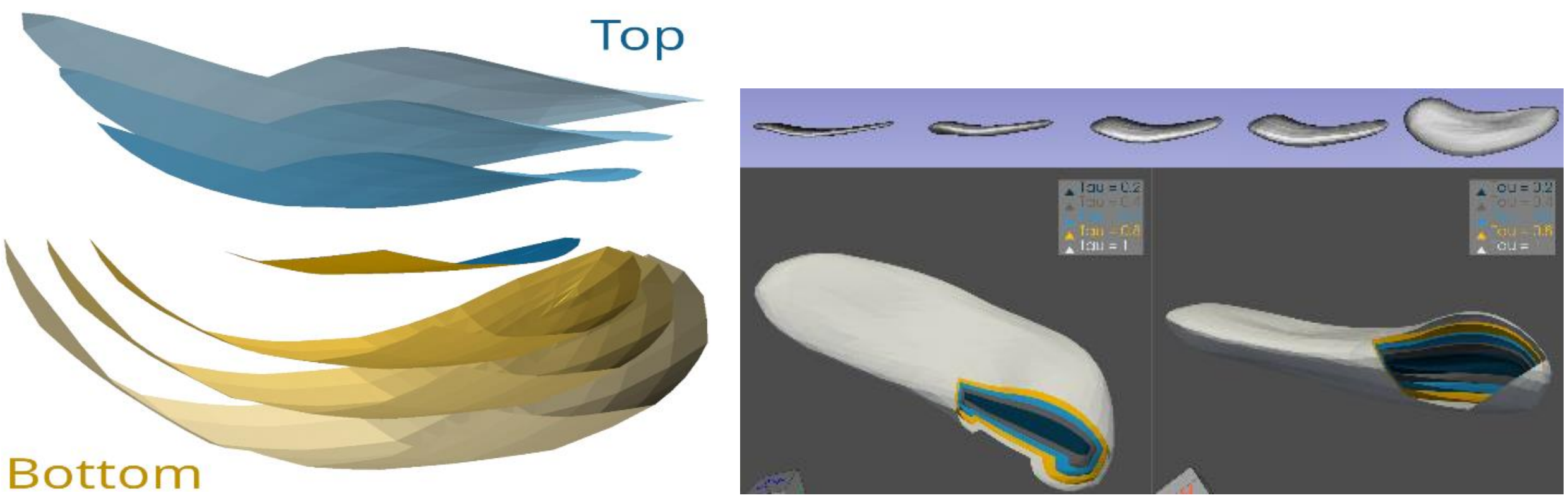

Fig. 7. Onion skins. Left: of a bent ellipsoid showing the top (blue) and bottom (yellow) portions of each of 4 onion skins presented using front-side, back-side, left-side and right-side clipping planes (and thus not showing where the skeleton folds). These onion skins are shown successively lighter as the skeleton (darkest; 0% of the way from the skeleton to the boundary, so both top and bottom are co-located), somewhat lighter (50% of the way), yet lighter (75% of the way), lightest (the boundary – all the way). Right: of a hippocampus with the top row showing 5 onion skins and the bottom row showing two views using a front clipping plane.

In the studies for this paper the fitted frames will be calculated along each spoke for $\tau_2$ (radial distance) values of 0.0 (the skeleton), ±0.25, ±0.5, ±0.75, and ±1.0 (the boundary position of the spoke), where the negative values refer to spokes on the bottom side. The normal to the onion skin there and the pure $\theta$ direction in the tangent plane to the onion skin are computed using small shifts of the relevant spokes, with the shifted spokes computed using the spoke interpolation described in [Liu 2021].

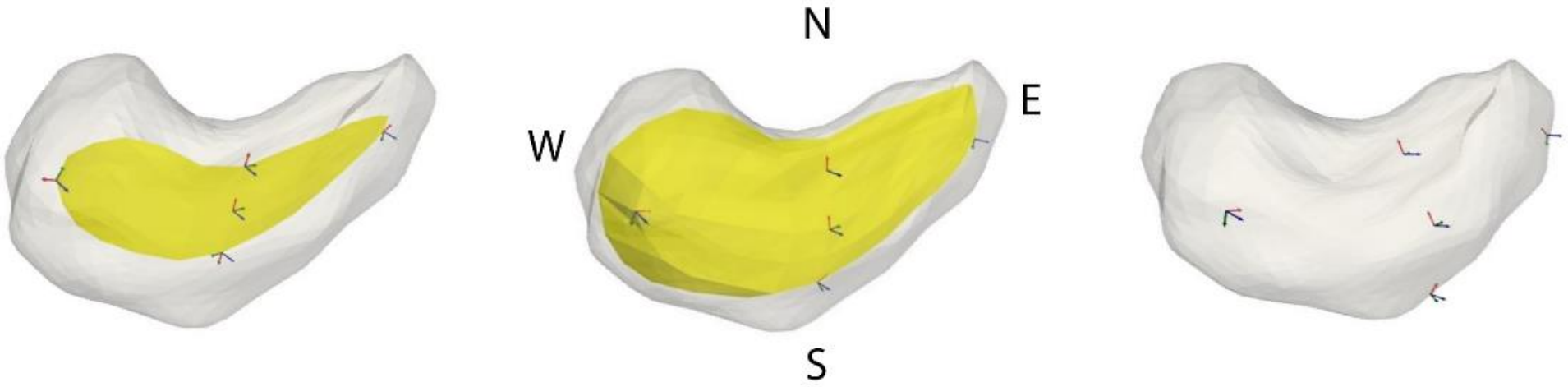

Fig. 8. Fitted frames at corresponding values of ($\theta$,$\tau_2$). Left: on the skeleton. Middle: on onion skin at $\tau_2$=0.5. Right: on the boundary

## 3. Geometric Features Derived from Fitted Frames

The geometric features that we will use to characterize any object are all derived from its s-rep. There are three types of features, which can be understood by reference to Fig. 4.: local frame curvatures of the onion skin at any onion skin point, local vectors describing positional shifts between onion skin points, and, at each skeleton point (with the topside and bottomside skeleton points taken as being different) the spoke vector and the frame curvature between the two ends of the spoke ($\tau_2$=0.0 and 1.0), taking care that the bottomside skeletal frames are rotations of the corresponding topside skeletal frames. Each of the features at an onion skin point are computed in the coordinate system provided by the frame at that point. The spoke-related features are computed in the coordinate system provided by the frame at the skeletal end of the spoke. The onion skin points used are all along the respective spokes. Special care is taken at the points on the skeletal fold and the interior points along their spokes, which have infinite curvature; this includes the skeletal vertices and their spokes' interior extensions.

Each onion skin curvature at a point on the skeleton (the $\tau_2$=0.0 onion skin), interior onion skin, or boundary (the $\tau_2$=1.0 onion skin) can be characterized by the rotation per unit distance between fitted frames at adjacent spokes one spoke away in positive and negative offsets in directions along the onion skin. At all but skeletal fold points and the interior onion skin points along the fold's spokes, we compute that for two roughly orthogonal directions $\theta$ and $\tau_1$; for skeletal fold points and interior onion skin points along the fold's spokes, where the curvature is infinite in the $\tau_1$ direction, we only compute curvature along the fold, i.e., the $\theta$ direction. For $\boldsymbol{v}$ either $\theta$ or $\tau_1$, the associated curvature at $\underline{x}(\theta, \tau_1, \tau_2)$ is computed by the rotation between the fitted frames at that onion skin's point that is $\Delta\boldsymbol{v}$ away from $(\theta, \tau_1, \tau_2)$ and that onion skin's point that is $-\Delta\boldsymbol{v}$ away from $(\theta, \tau_1, \tau_2)$, with the rotation rate about its axis divided by the Euclidean distance of $\underline{x}$ between those two onion skin points. As previously stated, those curvatures are given in the coordinate system of the fitted frame at $(\theta, \tau_1, \tau_2)$. So at each onion skin point other than the skeletal fold, there will be curvatures computed for two directions, with each curvature being a 3-tuple $\in$ SO(3); at the skeletal fold there is one curvature 3-tuple computed, along the fold. At the skeleton the curvatures are only computed for non-negative $\tau_2$, the top side of the folded skeleton.

Each positional shift at an onion skin point is characterized by the aforementioned inter-$\underline{x}$ vector between the two onion skin points $\pm\Delta \boldsymbol{v}$ away in s-rep coordinates. So at each onion skin point other than the skeletal fold, there will be positional shift vectors computed for two directions, with each positional shift vector being a 3-tuple $\in R^{+}\times S^2$; again only one positional shift vector is computed along the skeletal fold. The direction components are taken in the local frame at the tail of the vector.

The geometric features used in s-rep based representation of our hippocampi that take part in our comparison experiments are described in detail in the companion paper [Vicory 2025]. The hippocampi are spatially sampled with 61 interior skeletal points, of which 6 are along the spine ends' extension, and 24 skeletal fold points.

To form the final Euclidean features used in the classification, every direction vector is Euclideanized by Principal Nested Spheres applied to $S^2$; every curvature is Euclideanized by Principal Nested Spheres applied to $S^3$; and every length is Euclideanized by taking its logarithm and subtracting the average (over the training sample) of those logarithm values of the very length property in question.

This yields a tuple of size 8,076 forming the geometric feature vector. The respective elements of this feature vector are in the geometry-based correspondence that we have described.

**4. Diffeomorphisms of 3D Space via LDDMM**

The publicly available program called *Deformetrica* [Durrleman 2014] uses an approach called LDDMM (Large-Deformation Diffeomorphic Metric Mapping) that is designed to take a collection of boundary points (and/or curves or surfaces) in correspondence between a source object and a target object and to yield an energy-minimizing diffeomorphism over all of 3-space that carries the locations in the source object to their corresponding locations in the target object. The settings of that program allow the surface-to-surface and curve-to-curve mapping either to match specific surface or curve points with those in the other or to let the energy minimization choose which points match which. For our experiment measuring the classification effectiveness of LDDMM on boundary-to-boundary matching, the corresponding mesh vertices of the two surfaces were specified to match.

Deformetrica's momentum result is represented over an object-containing 3D Cartesian grid (see Fig. 9). It is typically preceded by a Procrustes alignment of the individual target objects in a training and test population. If an object summarizing the collection of aligned data objects, such as their Euclidean mean, is used as the source object, the result for each target object is an array surrounding the object, in which each element is a displacement vector capturing how the diffeomorphism initializes from the identity. The idea is that the displacement array's vectors, each in a global frame produced by alignment, carries the shape-descriptive geometric features that can be used in statistical operations such as classification. In 3D the array of N displacement vectors consists of 3N features. More details of the LDDMM methods used in this project are provided in the companion paper [Vicory 2025].

Application of Deformetrica to each object in the data set computes the diffeomorphism between the mean object and that object.

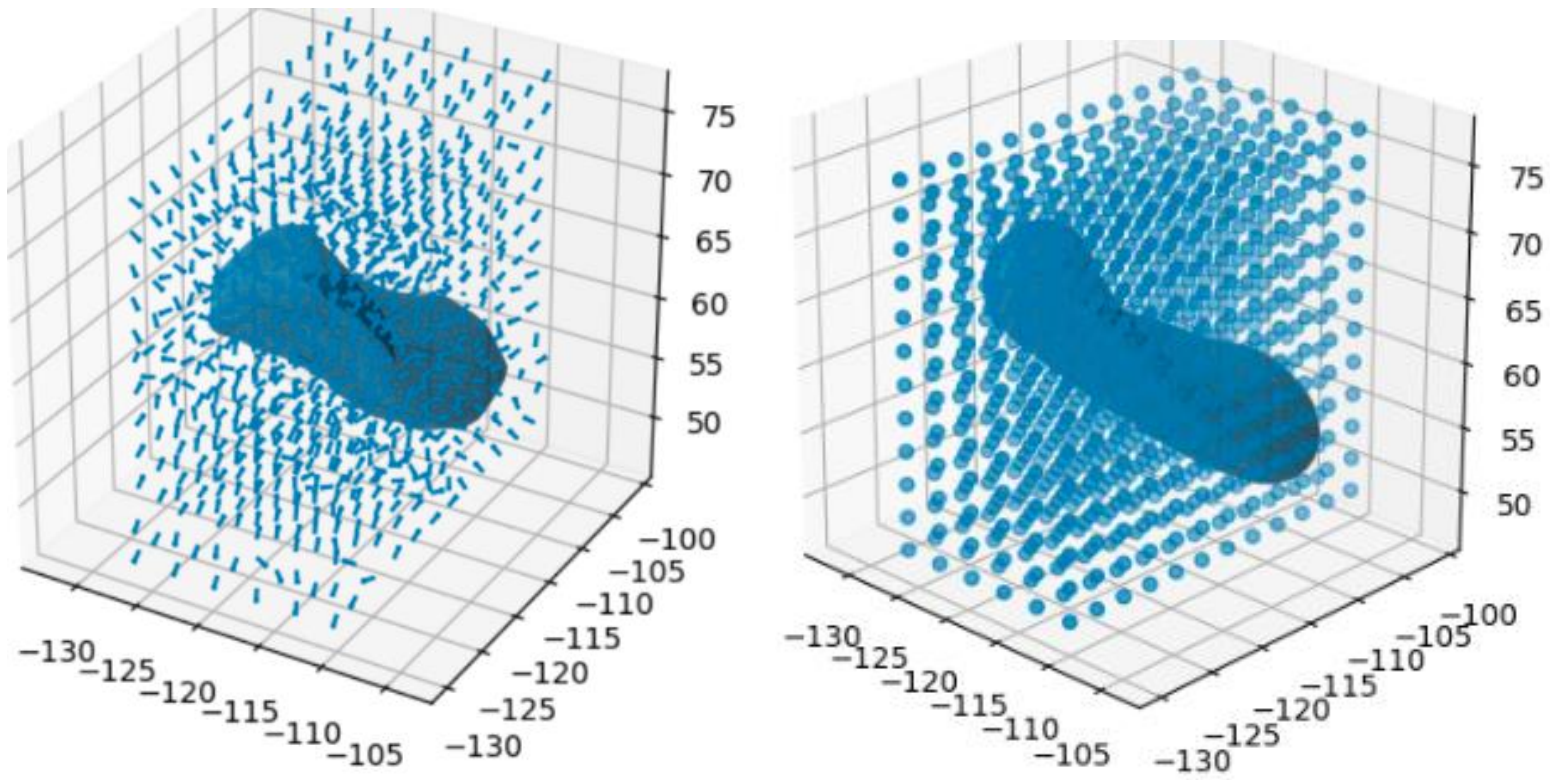

Fig. 9. Left: Mean hippocampus in grid of vectors. Right: A target hippocampus in the grid of vectors.

## 5. Fitting an S-rep to an Object Boundary

Both the Liu method and the Pizer/Vicory evolutionary method for fitting an s-rep to a target object operate on the principle that the input boundary mesh representing each object in the population can be smoothly mapped to a common ellipsoid and that the s-rep analytically derived from that ellipsoid can be diffeomorphically mapped back to the target object. For our target problem of anatomic object statistics, there is no issue of the discrete orientations of the common ellipsoid with its symmetries, as all of the objects in the population come from medical images in which the body-implied orientations are consistently the same

For the method to apply to any object of spherical topology, producing no singularities, we use the conformalized mean curvature flow method (CMCF) [Kazhdan 2012] for the forward flow. For each point in the target object, that forward flow yields a corresponding boundary point.

Very early in the CMCF mapping stages, whatever protrusions and indentations that were on the target object disappear. Soon thereafter, the curvatures of the object straighten out. See Fig. 10, where the mandible resolves into what we call a “bent hotdog”, followed by its straightening.

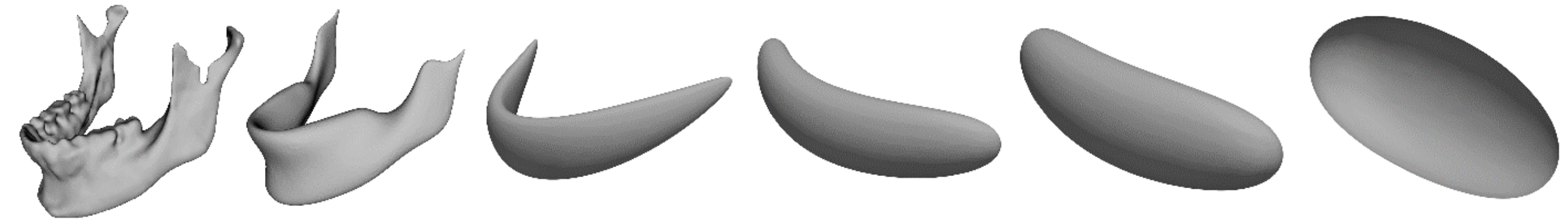

Fig. 10. Various stages of CMCF applied to a mandible. From left to right: the mandible, its protrusion-free host for its protrusions, the “bent hotdog”: showing its general curvature, the more straightened hotdog, the object nearing the ellipsoid, and the resulting ellipsoid.

For objects that do not have flat sections[5], CMCF yields a sphere in the limit, but it nears an ellipsoid on the way. And that ellipsoid can be diffeomorphically mapped in an approach respecting the s-rep properties [Damon 2021] to the mean of the training objects' ellipsoids.

The CMCF is applied in stages increasing from $t=0$ at the target object. At each stage of CMCF the boundary mesh computed is tested for nearness to an ellipsoid. Computing when the CMCF is near enough to an ellipsoid, i.e., determining when the flow is almost done, defining $t=T-1$, is done by fitting the flow-applied points to an ellipsoid using eigenanalysis of the matrix of second moments about the center of mass of the points. When the flow-applied points projected to the fitted ellipsoid have all short enough projection lengths, the flow is stopped, but for one final stage, which maps the near-ellipsoid to the actual ellipsoid, which we specify as being at stage $T$. This final step of the mapping from the target object to its best fitting ellipsoid is defined by the aforementioned projections. Over all training objects, we use a common threshold for closeness to its ellipsoid.

### *5.1 Liu method*

In the Liu method [Liu 2021] the points on the ellipsoid are mapped back to the points on the target object by a single diffeomorphism, which was computed via Thin Plate Splines (due to F. Bookstein but best described in [Dryden & Mardia 1998]). This diffeomorphism is then applied to the points on the discretely modeled ellipsoid's skeletal points and its corresponding spoke-implied boundary points to form vectors in the target object s-rep, i.e., its spokes. However, this result has four weaknesses: the implied skeletal points are more irregularly spaced than they need to be for correspondence across the training entities, the corresponding spokes at each skeletal point have lengths that are too different and such that their ends are less close to boundary than they might be for a good fit to the input object boundary interpolated from the object representation, and the spokes are farther from the medially implied objective of orthogonality to the skeletally implied boundary than they need to be.

To ameliorate these weaknesses, as described in [Liu 2021], a second stage modifies the s-rep to optimize measures of the properties of weakness. Fig. 11 shows the s-reps at each of the stages for a hippocampus.

The means of producing fitted frames, applying the refinement step, and calculation of the geometric features is exactly the same as with the s-reps from the evolutionary method.

[5] A special mapping to an ellipsoid for objects that do have flat sections has been designed by Vicory and Pizer.

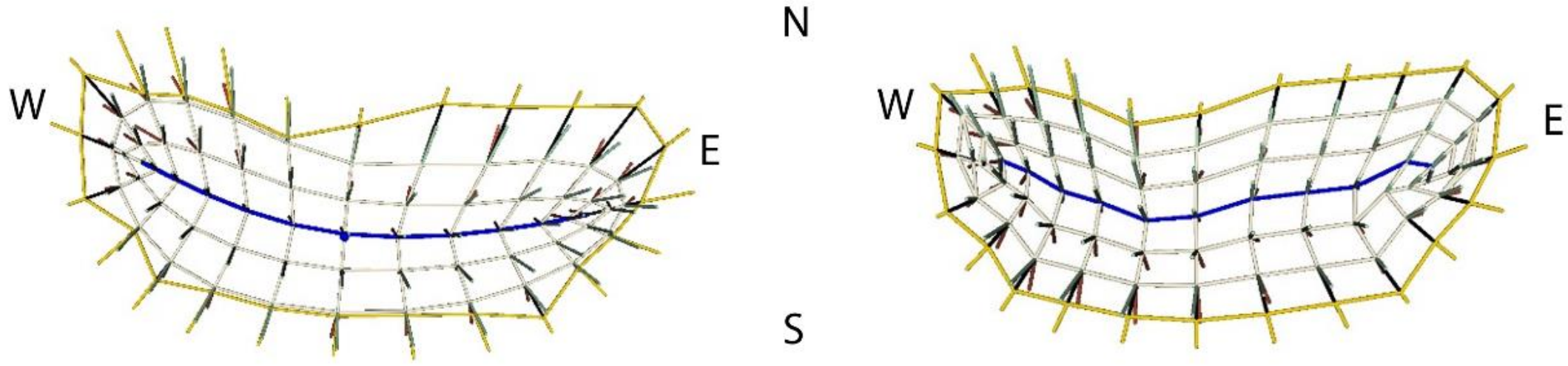

Figure 11. S-reps from Liu's method. Left: the first stage s-rep result. Right: the refinement stage's s-rep result**.**

### *5.2 The evolutionary s-reps method: a new method for s-rep fitting*

Vertices[6] and crests[7] on a surface are geometrically important loci. From the skeletal point of view, crests are surface loci to which spokes from the folds of the skeleton proceed, and vertices are the surface points to which spokes from vertices of the fold proceed. That is, they determine the bounding points and curves of the skeleton. The steps of CMCF do not map these geometrically important loci onto corresponding loci at any of its stages. However, we wish the s-rep to evolve at each of the stages of the backward mapping (deriving a diffeomorphism) from the ellipsoid at $t=T$ to the target object at $t=0$, while respecting the aforementioned properties of the skeleton at those fold and vertex points. That is, these points and curve must map back from the ellipsoid's points and curve to the corresponding points and curve on the successive surfaces produced from the forward flow, and eventually to the target object. Also, the centers of curvatures of the vertices, i.e., the skeleton vertices, should map to each other across the stages. In addition, the centers of curvatures of the crest curves, which are the fold curves of the skeleton, should map to each other across the stages.

But even this is does not capture enough of the object geometry. The s-reps themselves should map back to corresponding s-reps associated with the surfaces associated with the stages, which came from the forward mapping. More precisely, the diffeomorphism from stage $t$ to stage $t-1$ should map the straight spokes at stage $t$ onto straight spokes at stage $t-1$, and even more strongly, it should map the radial distances on each of those spokes at stage $t$ onto the corresponding radial distances at stage $t-1$.

Thus, the diffeomorphism defining the backward flow (back to the target object from $t=T$ to $t=0$) needs to respect the surfaces produced at various values of $t$, but it has to respect all of the aforementioned geometric constraints as it computes this backward flow in a series of stages. Then the overall diffeomorphism from the ellipsoid to the target object is produced by composing the inter-stage diffeomorphisms.

Fig. 12 shows how the Vicory [2025] program accomplishes the s-rep calculation from its value at stage $t$ to its value at stage $t-1$.

---

[6] A vertex is a local relative maximum of convex Gaussian curvature.
[7] A crest point is a local relative maximum along one of the principal directions of one of the principal curvatures.

At each of the stages, the program has the following pairs between stage *t* and stage *t-1*:

1) the object surfaces;
2) computed by pre-processing, the crest curves (the ones that were associated with those of the ellipsoid) and the corresponding skeletal fold curves;
3) computed by pre-processing, the vertices (the ones that were associated with those of the ellipsoid) and the corresponding skeletal vertices.

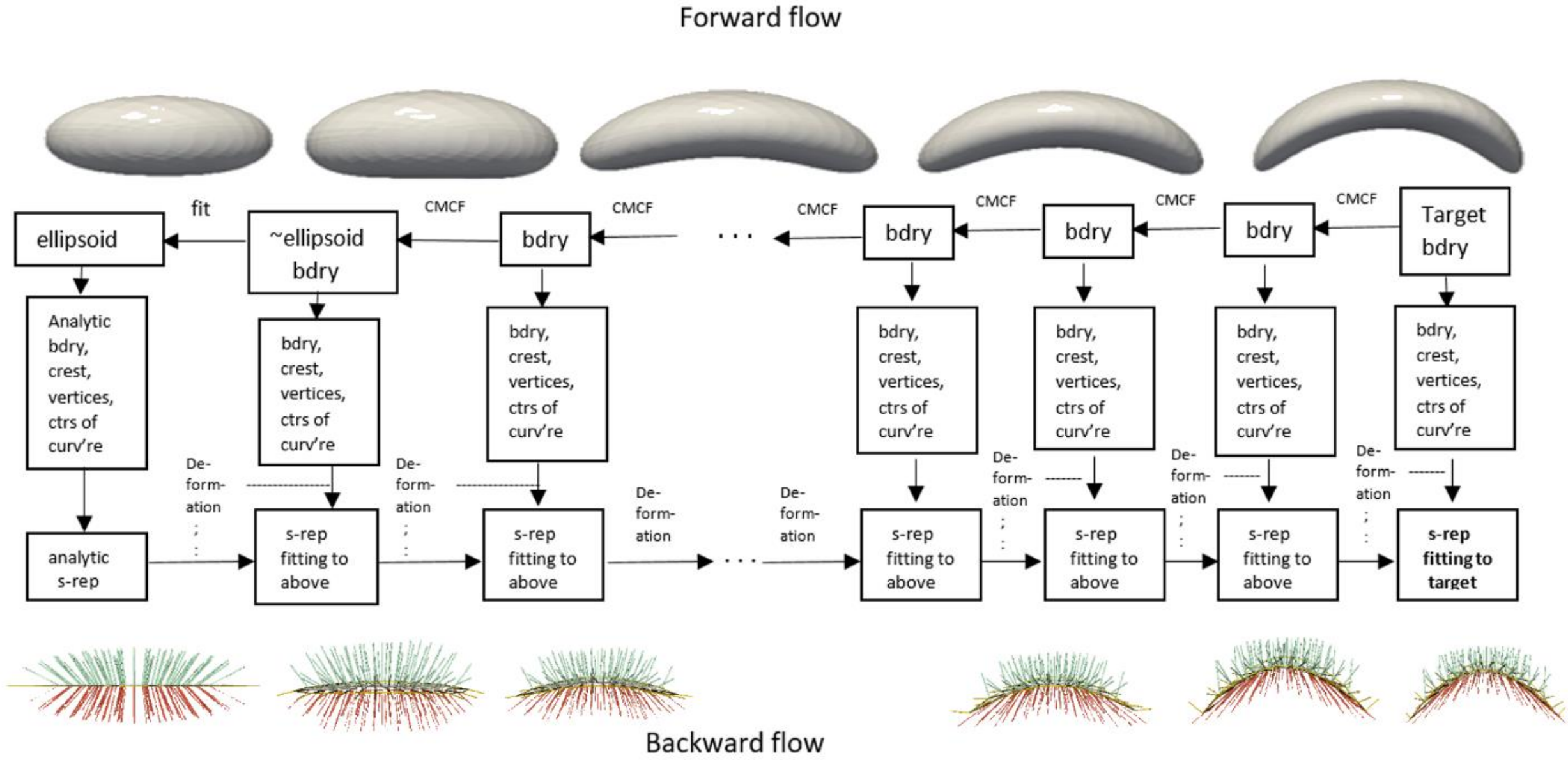

Fig. 12. The s-rep fitting method on a bent ellipsoid. Right to left is forward (CMCF) flow. Left to right is the backward (s-rep fitting) flow, from the ellipsoid to the target bent ellipsoid.

Using those, with no specific point-to-corresponding-point matching in the surfaces and crests, Deformetrica is used to compute an initial diffeomorphism from stage *t* to stage *t-1*. That diffeomorphism is applied to the s-rep at stage *t* to produce an initial approximation to the s-rep at stage *t-1*.

However, just as with the Liu method (which is applied only to the global backward flow transformation) this initial approximation for a stage needs improvement to make it a better skeletal representation of the object surface at stage *t-1*. Its skeletal positions need to be regularized, both along the fold and within the skeletal grid. Its spokes need to become closer to orthogonal to the object boundary, and the skeletal ends shared by spoke pairs need to move in the direction of the vector between their boundary ends to become approximately of equal length. In producing near orthogonality, crossing of adjacent spokes must be avoided. And the resulting skeletal surface must be smooth.

Once these modifications have been made, a new diffeomorphism between stage *t* and stage *t-1* can be computed using Deformetrica. This diffeomorphism would respect not only the correspondences used to produce the initial inter-stage diffeomorphism for that stage but also correspondences between the corresponding s-reps' spokes points of equally spaced radial distance, from 0 to 1. We have used radial distance spacing of ¼; we have used the skeletal spine spacing of 12 steps; and we have used vein spacing (crest to crest) of 7 steps.

Fig. 13 shows examples of the s-reps from both the evolutionary method, both for a simulated object formed as a bent ellipsoid and for one of the hippocampus samples.

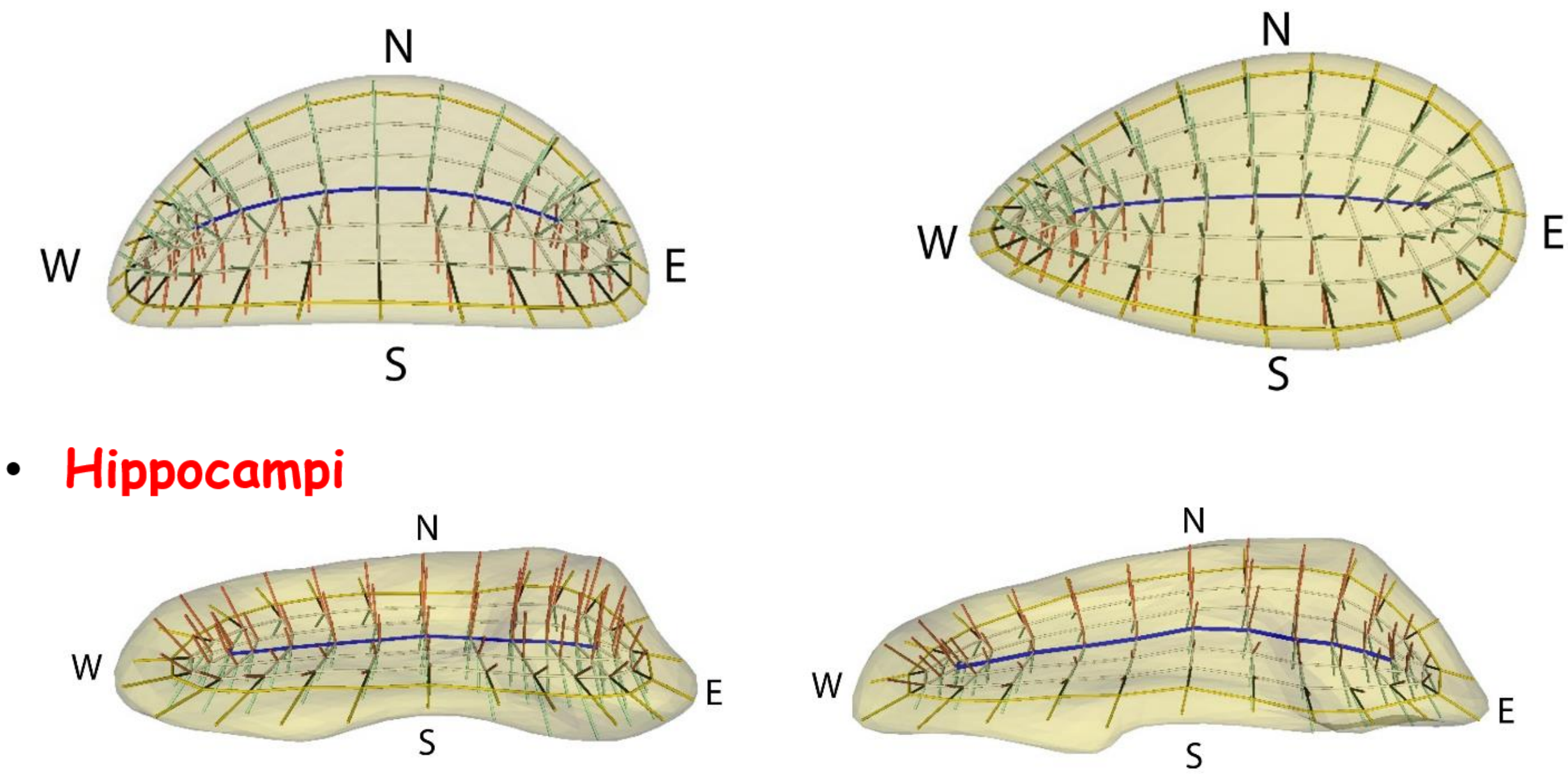

Figure 13. S-rep fitting results from the Pizer/Vicory evolutionary method.

## 6. Testing via Classification Performance

### *6.1 Materials*

The objects with which we compare the classification performance of our three methods are hippocampi of 6-month olds (see Fig. 4) in two classes: those that developed autism symptoms in later years and those that did not. This challenging classification problem is to predict from a hippocampus whether the young patient will later develop autism. Previous work using among others, cortical shape features, suggested that diagnosis (classification) of impending autism via subtle brain shape features might be possible [Hazlett 2017]. We chose the hippocampus object because autism is known to affect that structure but previous methods to classify it barely passed pure guessing in accuracy, with AUCs at most 0.6 (see Table 1), and clinical experts could not sense the differences. In that data set there are 34 cases in which autism developed and 143 in which it did not.

Each hippocampus was segmented from a registered pair of T1-weighted and T2-weighted MRI images and is data produced by the IBIS consortium. The meshes forming the data for our methods were produced using spherical harmonic analysis [Styner 2006]. The s-reps we used were made of 61 interior skeletal points and 24 skeletal fold points. This leads to a feature tuple of size 8,076 representing a hippocampus. All of these features have been Euclideanized. This large number of 8,076 Euclideanized features has been reduced by PCA to 1 less than the number of data samples: 177.

For the diffeomorphism method, we have chosen to have a number of starting features from each sample to also be much larger than the number of training samples. In particular, we have chosen to represent the diffeomorphism in a momentum grid (or initial velocity grid, see below) of $6 \times 6 \times 12$ with there being a vector at each grid point. The result is a feature tuple of size $6 \times 6 \times 12 \times 3 = 1296$, to which Euclideanization has been applied. Again, this set is notably larger than 1 less than the number of data samples: 177, a number to which the 1296 features have been reduced by PCA.

Two forms of the LDDMM method are evaluated. The first uses Deformetrica between the mean object and the target object represented as boundary meshes and yields a momentum grid. The second uses a program due to Zhang [2019] between the mean object and the target object represented as binary images and yields a grid of initial velocity vectors. In both forms the momenta, resp. the initial velocities on deformations from the mean object to the target object were used, after Euclideanization, as the geometric features.

However, to avoid a statistically invalid evaluation where the full data is used in forming the geometric features, for the LDDMM method the mean was computed over a random quarter of both classes of the data and the evaluation was produced by the random holdout method on the remaining ¾ of the data. For each of the 1000 times the AUC and AUPR were computed, this process was repeated 6 times and the resulting AUC and AUPR measures (see below) were averaged over the repetitions. We checked and verified that the results were not materially affected by this form of averaging by repeating the experiment with the standard approach in the literature in which the mean is computed from the whole data set and the random holdouts are applied to the whole data set.

### *6.2 Methods*

We will compare our four representations, LDDMM (in two forms) vs. Liu's s-reps vs. our evolutionary s-reps, in the following way. We will be comparing representations each of which has 177 features. Each of its derived representations is used in a classification method to yield an Area Under an ROC Curve (AUC) as well as an Area Under the Precision, Recall Curve (AUPR), on the basis of which the success of each of the three methods can be judged. The AUC is a commonly used measure; the AUPR was chosen due to its lower sensitivity to the imbalance between the number of cases of each class and its greater focus on successfully classifying the positive cases.

The method of random holdouts was applied using the classification method entitled Distance Weighted Discrimination (DWD) [Marron 2007]. For each holdout, the non-held-out cases are used to produce a separation direction in feature space, upon which all of the training cases are projected, forming two histograms, one for each class. Using those histograms, the public software for the method called SMOTE [Chawla 2002] was used to form an AUC and an AUPR for the collection of holdouts [Liu 2021]. The AUCs over 1000 holdouts produce the overall AUC and AUPR for that object representation method.

### *6.3 Results*

Table 1 gives the AUC values for each of the four object representations.

While it is not possible to attach significance levels to these AUC and AUPR values, due to the high correlation of the various random holdouts, the results indicate that the notably superior method for producing a representation aimed for statistical analysis is our evolutionary s-rep fitting method proposed in this paper. Of course, this result is on only one anatomic structure with respect to only one mental disorder, but it nevertheless suggests the superiority of representing the object interior using fitted frames, as well as the superiority of fitting s-reps to object boundary meshes by deformation of their interior closure from an ellipsoid in a way recognizing maintenance of s-rep relevant properties throughout the deformation.

| **Representation** | **AUC** | **AUPR** |
|---|---|---|
| Evolutionary s-reps | **0.73** | **0.38** |
| Liu s-reps | 0.60 | 0.34 |
| Mesh diffeomorphism momenta | 0.53 | 0.23 |
| Binary diffeomorphism initial velocities | 0.58 | 0.27 |

Table 1. AUCs and AUPRs for the Pizer/Vicory s-reps, Liu s-reps, and diffeomorphism momenta and initial velocities, respectively. For both measures, a larger value shows better classification performance.

## 7. Producing Non-self-intersecting Object Means

Object means are necessary for a variety of statistical operations, including hypothesis testing by permutation approaches. The concern is that these means be guaranteed to be geometrically valid. In particular, anatomic objects do not self-intersect. Yet it seems that every method of computing object means, including the evolutionary s-reps method described in this paper, can produce local self-intersections even when no members of the population whose mean is being computed has self-intersections. Assuming that the mean is computed by the Fréchet approach of minimizing the sum of squared geodesic distances in the shape space from the population entities to the candidate mean, the need is for that shape space to include no self-intersecting objects and for the geodesic paths thus to pass only through non-self-intersecting objects. This property can be guaranteed if in a shape space of objects represented by swept surfaces, such as s-reps, every object satisfies Damon's [2021] relative curvature criterion (RCC). This criterion specifies that at each point on the spine, 1.0 > the product of the curvature of the spine and the distance to the closest point on the cross-section boundary from the spine.

Taheri [2025] has shown how to provide such a shape space for a particular category of s-reps. These simple s-reps, which he calls "Elliptical Tube Representations" (ETReps) are ones where the cross-sections have the form of a planar ellipse orthogonal to the spine's tangent (see Fig. 14 for examples). We leave the details of that method to the references, but we note the possibility that this type of shape space could be achieved for more general s-reps.

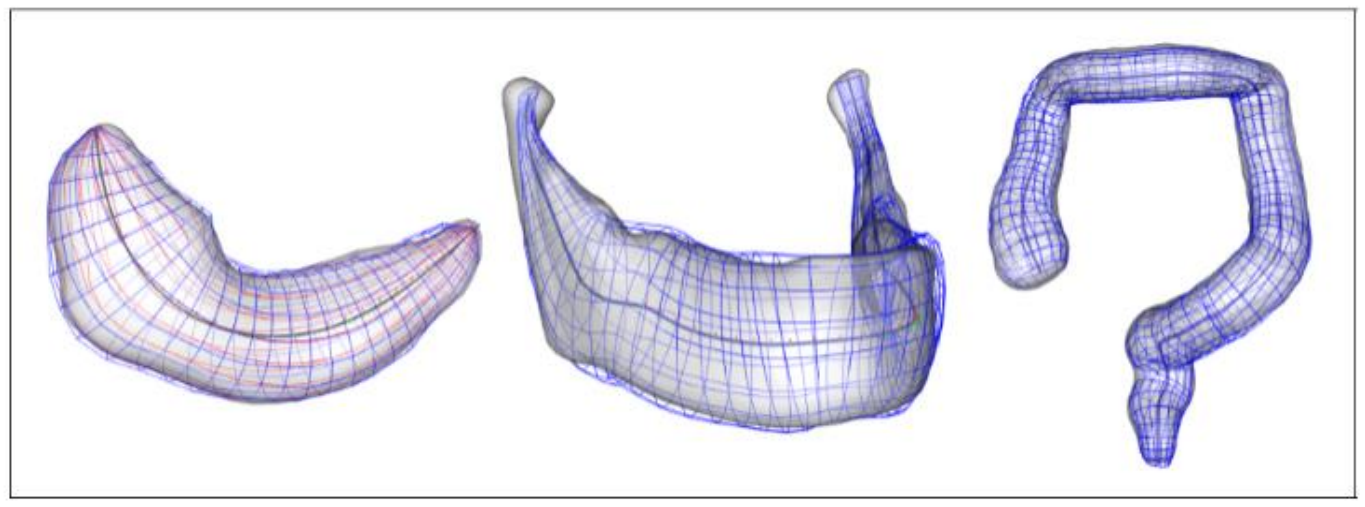

Fig, 14. Elliptical Tube Representations (ETreps) for three anatomic objects. Left: a hippocampus; Right: a colon, Center: A mandible smoothed by CMCF to remove its protrusions.

## 8. Conclusions

This paper described a novel means of generating *fitted frames* in the closure of *an object's interior* and then generating alignment-independent geometric curvature and positional spacing primitives from those frames. This appears intellectually to be a notable advance, since in this application and just a few others [Pizer 2022, Liu 2023, Taheri 2023] it has shown its strength for generating statistically useful shape features, not only locally but across inter-object and intra-object locations.

Besides the mathematical novelties of fitted frames in an object's interior, our approach producing correspondence across a population of object instances is methodologically novel. Instead of deriving zero-order (positional) or first-order (deformational vectors) geometric features directly from the object boundary, it fits a model of the closure of the object interior such that the fitting itself is nonlinearly, evolutionarily based on a rich variety of geometric properties that presumably are shared across the population. And instead of using the s-rep model's features directly in the statistics, it derives vector and curvature features from the s-rep's fitted frames.

The correspondence comes from the fact that the evolution of each object in the population starts from a common ellipsoid, with each evolutionary stage producing an optimized deformation into the object in a fashion based on shared geometric properties. The resulting s-rep model, namely its spine, veins, and spokes, is the descriptor of the target object. Some ask for what the mathematical descriptor of these entities is relative to the object boundary. Instead, one must realize that those entities are defined algorithmically. For example, the spine, which conceptually is a skeletal descriptor of the object's skeletal surface, is defined specifically as the result of the evolutionary deformation of the ellipsoid's spine, the latter being mathematically known.

The results of our evaluation confirm this behavior via its superior performance in classification based on shape, albeit in only one population of anatomic objects. In particular, it suggests that an object representation that highly recognizes the shape properties of the object interior, not just globally within the object but locally as well, can produce better statistics than one that is based on the limited properties of the object boundary alone.

The program for s-rep fitting made a variety of decisions that might later be upgraded. Moreover, the program runs rather slowly: the present speed of the program on salt.slicer.org that fits an s-rep to a boundary mesh is ~5.5 minutes, mostly taken up by the many applications of Deformetrica. However, programs have been developed [Gaggion 2024] to fit similar skeletal models to objects' boundary meshes using a deep learner that in application runs in around 1 second. In the near future such programs, trained from the present program, will be producible from the training set for any object.

The details of the s-rep fitting method and of the comparison experiment are provided in the companion paper [Vicory 2025]. The software underlying the method [Vicory 2018, 2023], including tutorials on its use, is available at the Slicer/SALT website: https://salt.slicer.org.

## 9. Extensions

This paper has dealt only with objects that are well characterized by a simple skeleton. It has not dealt with the more complicated shapes that have statistically salient protrusions or indentations (we call those subfigures) that might need their own skeleton, nor with multi-object complexes. As the CMCF forward flow that was applied causes the subfigures in a multifigure complex to melt away into the host figures, it appears possible to detect this behavior and represent the subfigure and its connection to the host figure skeletally. This approach would allow the methodology of s-rep fitting to be applicable to many more objects.

In addition, the method given here need not be restricted to objects with spherical topology. For example, extension to objects with toroidal topology might be developed.

As for multi-object statistics, Liu [2023] has shown how the within-object correspondences in such complexes can be extended to descriptions of the inter-object geometry, at least for objects that are somewhat separated and roughly parallel. There is much promise in extending that work to other multi-object arrangements.

Finally, we return to the matter that motivated the geometric representation proposed here, namely correspondence. This is an attractive, open area for further research. Here the correspondence was achieved via only geometric features, but there are plenty to consider as to other features to add over the features of object interiors (and in other work, inter-object region geometric features) considered here. When the objects come from images, correspondence on image intensity features or derived texture features, in object-based coordinates such as the s-rep coordinate system presented here, can be important and could be appended to the shape-based features in producing the correspondences in the backward flow. Other location-based features that depend on the driving area in which the objects fall, such as biological features in biomedical problem areas, could also be important.

## Acknowledgements

We appreciate help on this project and/or in writing this paper from James Fishbaugh, Md Asadullah Turja, David Allemang, and Ankur Sharma.

Funding: This work was done with the partial support of NIH grant R01EB021391.

## References

Cartan, E. La structure des groupes de transformations continus et la théorie du trièdre mobile, Bull. Soc. Math. France, t. 34: 250-284. 1910.

Cates, J., Fletcher, P. T., Styner, M., Shenton, M., & Whitaker, R. Shape modeling and analysis with entropy-based particle systems. In Information Processing in Medical Imaging: 20th International Conference, IPMI 2007, Kerkrade, The Netherlands, July 2-6, 2007. Proceedings 20 (pp. 333-345). Springer Berlin Heidelberg. 2007.

Chawla, NV et al. "SMOTE: synthetic minority over-sampling technique." Journal of Artificial Intelligence Research 16: 321-357. 2002

Cootes, T F, & Taylor, CJ. Statistical models of appearance for medical image analysis and computer vision. In Medical Imaging 2001: Image Processing (Vol. 4322, pp. 236-248). SPIE. 2001

Cury, C, JA Glaunès, R Toro, M Chupin, G Schumann, V Frouin, J-P Poline, O Colliot, and the Imagen Consortium. Statistical Shape Analysis of Large Datasets Based on Diffeomorphic Iterative Centroids. *Front. Neurosci.*, 12 November 2018, Sec. Brain Imaging Methods, Volume 12 - 2018 | https://doi.org/10.3389/fnins.2018.00803

Damon, JN. Geometry and medial structure. Ch. 3 in [Siddiqi 2008] 2008a

Damon, J. "Swept regions and surfaces: Modeling and volumetric properties." Theoretical Computer Science 392.1-3 (2008b): 66-91.

Damon, JN. Thoughts on Ellipsoidal Models. Personal Communication. 2021

Dryden IL, KV Mardia. *Statistical Shape Analysis.* Wiley, 1998. Also available in a second edition, 2016.

Durrleman, S., Prastawa, M., Charon, N., Korenberg, J. R., Joshi, S., and Gerig, G. Morphometry of anatomic shape complexes with dense deformations and sparse parameters. Neuroimage 101, 35–49. doi: 10.1016/j.neuroimage.2014.06.043. 2014

Gaggion, N, E Ferrante, B Paniagua, J Vicory. Fitting Skeletal Models via Graph-Based Learning." In *2024 IEEE International Symposium on Biomedical Imaging (ISBI)*, pp. 1-4. IEEE, 2024.

Hazlett, H. C. *et al.* Early brain development in infants at high risk for autism spectrum disorder. *Nature* **542**, 348–351. 2017.

Hong, JP, J Vicory, J Schulz, M Styner, JS Marron, SM.Pizer. Non-Eudlidean classification of medically imaged objects via s-reps. *Medical Image Analysis*, no. 31, pp. 37-45, 2016.

Kazhdan, M, J Solomon, M Ben-Chen. Can mean-curvature flow be modified to be non-singular? *Comput. Graphics Forum* 31, 1745–1754. doi: 10.1111/j.1467-8659. 2012.03179.x. 2012

Koenderink, JJ. *Solid Shape.* MIT Press, 1990.

Liu, Z, JP Hong, J Vicory, JN Damon, SM Pizer. Fitting unbranching skeletal structures to objects. *Medical Image Analysis*. 2021.

Liu, Z, J Damon, JS Marron, S Pizer. Geometric and Statistical Models for Analysis of Two-Object Complexes. *Int. J. Comp. Vis.* 1126-1123, 2023

Marron, JS, MJ Todd, J Ahn. Distance weighted discrimination. *Journal of the American Statistical Association*, 102 (480): 1267–1271. 2007.

Pizer, SM and JS Marron. Objects statistics on curved manifolds. In *Statistical Shape and Deformation Analysis*, no. (G. Zheng, S. Li, and others, eds.), 2017.

Pizer, SM, JS Marrron, JN Damon, J Vicory, A Krishna, Z Liu, M Taheri. Skeletons, Object Shape, Statistics. *Frontiers in Computer Science*, 18 October 2022, Sec. Computer Vision https://doi.org/10.3389/fcomp.2022.842637, 2022.

Styner, M., et al. (2006, 01). Statistical shape analysis of brain structures using SPHARMPDM. The insight journal, 1071 , 242-250. 2026.

Taheri, M, SM Pizer, J Schulz et al. Fitting the discrete swept skeletal representation to slabular objects. Under review for journal publication. PREPRINT, Research Square [https://doi.org/10.21203/rs.3.rs-2927062/v1]. 2023.

Taheri, M., Pizer, S. M., & Schulz, J. The Mean Shape under the Relative Curvature Condition. To appear in the Journal of Computational and Graphical Statistics. 1-21. 2025. https://doi.org/10.1080/10618600.2025.2535600.

Tu, L; M Styner; J Vicory, S Elhabian; B Paniagua; JC Prieto; Dan Yang; R Whitaker; SM Pizer. Skeletal Shape Correspondence through Entropy. *IEEE Trans. Med. Img.*, **37**: 1-11, 2016

Vicory J, Pascal L, Hernandez P, Fishbaugh J, Prieto J, Mostapha M, Huang C, Shah H, Hong J, Liu Z, Michoud L, Fillion-Robin JC, Gerig G, Zhu H, Pizer SM, Styner M, Paniagua B. SlicerSALT: Shape AnaLysis Toolbox. Shape Med Imaging. 2018 Sep;11167:65-72. doi: 10.1007/978-3-030-04747-4_6. Epub 2018 Nov 23. PMID: 31032495; PMCID: PMC6482453. 2018.

Vicory J, Han Y, Prieto JC, Allemang D, Leclercq M, Bowley C, Scheirich H, Fillion-Robin JC, Pizer SM, Fishbaugh J, Gerig G, Styner M, Paniagua B. SlicerSALT: From Medical Images to Quantitative Insights of Anatomy. Shape Med Imaging. 2023 Oct;14350:201-210. doi:

10.1007/978-3-031-46914-5_16. Epub 2023 Oct 31. PMID: 38250732; PMCID: PMC10798161. 2023.

Vicory, J, N. Tapp-Hughes, J Zhang, Z Liu, SM Pizer. Fitting, Extraction, and Evaluation of Geometric Features To Produce Correspondence for Statistics. Companion paper to this paper. 2025.

Zhang, M. and Fletcher, P.T., 2019. Fast diffeomorphic image registration via fourier-approximated lie algebras. *International Journal of Computer Vision*, *127*, pp.61-73.